\pdfoutput=1

\documentclass[11pt]{article}

\usepackage{EMNLP2023}

\usepackage{times, bbm}
\usepackage{latexsym}
\usepackage[utf8]{inputenc}

\usepackage[T1]{fontenc}

\usepackage[utf8]{inputenc}

\usepackage{microtype}

\usepackage{inconsolata}
\usepackage{listings}   
\usepackage{graphicx}
\graphicspath{ {./images/} }
\usepackage{multirow}
\usepackage{amsmath,amssymb,amsfonts}
\usepackage{booktabs}
\usepackage{algorithm}
\usepackage{algorithmicx}
\usepackage{algpseudocode}
\usepackage[utf8]{inputenc}
\DeclareUnicodeCharacter{03C3}{\ensuremath{\sigma}}

\title{Automated Evidence Extraction and Scoring for Corporate Climate Policy Engagement: A Multilingual RAG Approach}
\author{Imene Kolli$^{1*}$ \quad 
        Ario Saeid Vaghefi$^{1*}$ \quad 
        Chiara Colesanti Senni$^{1*}$ \\ \bf
        Shantam Raj$^{2}$ \quad 
        Markus Leippold$^{1,3}$ \\\\
        $^1$University of Zurich, Department of Finance \\
        $^2$University of Zurich, Department of Informatics \\
        $^3$Swiss Finance Institute \\\\
        \thanks{$^*$Equal contribution.}
}

\begin{document}
\maketitle
\begin{abstract}
InfluenceMap's LobbyMap Platform monitors the climate policy engagement of over 500 companies and 250 industry associations, assessing each entity's support or opposition to science-based policy pathways for achieving the Paris Agreement's goal of limiting global warming to 1.5°C. Although InfluenceMap has made progress with automating key elements of the analytical workflow, a significant portion of the assessment remains manual, making it time- and labor-intensive and susceptible to human error. We propose an AI-assisted framework to accelerate the monitoring of corporate climate policy engagement by leveraging Retrieval-Augmented Generation to automate the most time-intensive extraction of relevant evidence from large-scale textual data. 
Our evaluation shows that a combination of layout-aware parsing, the Nomic embedding model, and few-shot prompting strategies yields the best performance in extracting and classifying evidence from multilingual corporate documents. We conclude that while the automated RAG system effectively accelerates evidence extraction, the nuanced nature of the analysis necessitates a human-in-the-loop approach where the technology augments, rather than replaces, expert judgment to ensure accuracy.
\end{abstract}

\section{Introduction}
Corporations significantly influence climate policy through lobbying activities, yet there is limited structured, accessible data capturing the nature and stance of this engagement \cite{leippold2024corporate}. Traditional lobbying registries and voluntary disclosures offer partial insight, often lacking clarity on whether corporate actions support or obstruct climate policy. InfluenceMap's \textit{LobbyMap} platform \footnote{\url{https://lobbymap.org/LobbyMapScores}} has been at the forefront of this effort, producing systematic evaluations of over 500 companies and 250 industry associations. However, much of this analysis remains manual, limiting both scalability and responsiveness to new disclosures.

We present \textbf{LobbyMap Search}, a multilingual RAG system that retrieves and classifies textual evidence of corporate climate policy engagement. Leveraging a pipeline of semantic search, LLM-based stance classification, and robust evaluation metrics, it allows stakeholders to audit lobbying behavior across thousands of corporate documents. This system provides a scalable method for quantifying climate lobbying, enabling comparative analyses across firms, sectors, and regions.

Our contributions include:
\begin{itemize}
    \item A novel retrieval-augmented stance classification pipeline tailored to climate lobbying, based on the InfluenceMap schema.
    \item Support for multilingual and multi-format corporate reports.
    \item Empirical evaluation framework, including both standard retrieval/classification metrics and recent oracle-based diagnostic scores.
    \item Detailed analysis of individual component performance and overall pipeline accuracy across four stance generation strategies. 
\end{itemize}

\section{System Overview}

LobbyMap Search processes company documents (PDFs) tagged with metadata (company, language, region) and computes lobbying stance scores for a predefined set of climate policy queries (see Appendix \ref{sec:queries}). The pipeline (Figure~\ref{fig:pipeline}) consists of five stages:

\begin{figure*}[t]
    \centering
    \includegraphics[width=\linewidth]{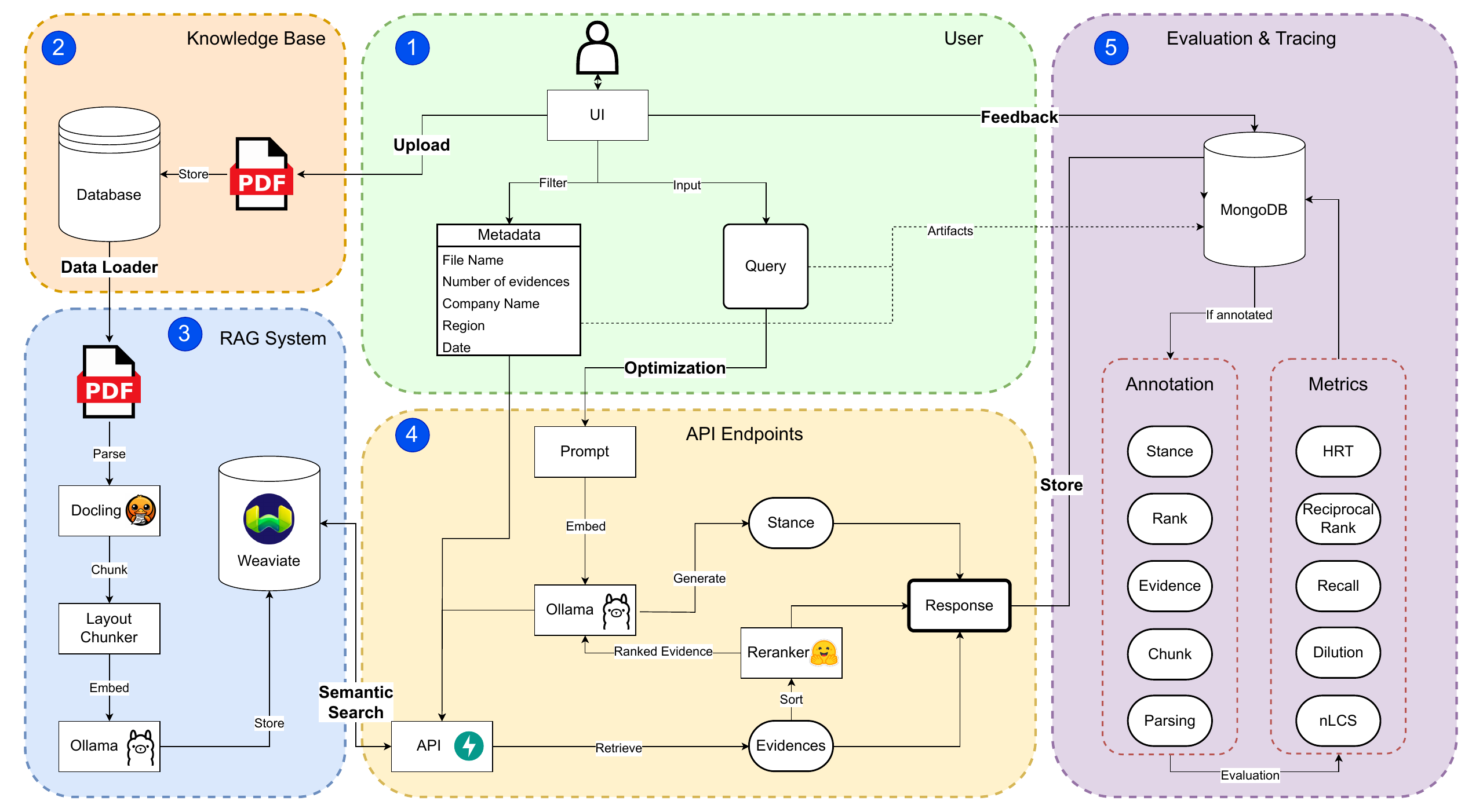}
    \caption{LobbyMap Search system pipeline.}
    \label{fig:pipeline}
\end{figure*}
\begin{enumerate}
    \item \textbf{User Interface}: Enables document uploads, query submissions, and feedback collection.
    \item \textbf{Knowledge base}: Stores uploaded documents and associated metadata for retrieval and reuse.
    \item \textbf{RAG system}: Parses and chunks documents, embeds and stores them in a vector database.
    \item \textbf{API Endpoints}: Handle user queries, return retrieved evidence, and facilitate interaction between components.
    \item \textbf{Evaluation and Tracking}: Logs user feedback, query artifacts, and evaluation results.
\end{enumerate}


\section{RAG Experimental Setup}
\subsection{Parsing and Chunking}
\paragraph{Parsing.} We use both Docling and PyMuPDF to parse PDF documents and compare their outputs across structure preservation and text quality.

\textbf{Docling}~\cite{livathinos2025doclingefficientopensourcetoolkit}, a layout-aware parser that outputs structured Markdown with headers, paragraphs, and tables. It uses EasyOCR for multilingual extraction.

\textbf{PyMuPDF}~\cite{pymupdf} offers faster multilingual linear parsing but lacks structural awareness; text is extracted as plain lines.

\paragraph{Chunking.} To enhance retrieval precision \cite{qu2024semanticchunkingworthcomputational}, we segment documents into coherent chunks:

\textbf{Semantic chunking}~\cite{chonkie2025} groups sentences based on embedding similarity.

\textbf{Layout-based chunking} uses structural markers and heuristic rules to define chunk boundaries.

\noindent A detailed configuration is in Appendix \ref{appendix:parser_chunker_config}.

\subsection{Embedding}

To support multilingual semantic retrieval, we embed text chunks into vector space using dense language models. We prioritize models with high language coverage and long-context capabilities to accommodate the diversity and length of corporate reports. We compare three models: \texttt{bge-m3} \cite{chen2024bgem3embeddingmultilingualmultifunctionality}, \texttt{nomic-embed-text:v1.5} \cite{nussbaum2024nomic}, and \texttt{Qwen3-Embedding-0.6B} \cite{zhang2025qwen3embeddingadvancingtext}. Model selection is based on retrieval quality, ensuring multilingual performance across document types.

\subsection{Reranking}

To provide users with the most relevant information first, we optionally apply reranking using models trained to predict relevance scores for chunk-query pairs. We experiment with \texttt{bge-reranker-v2-m3} \cite{chen2024bgem3embeddingmultilingualmultifunctionality}, \texttt{mxbai-rerank-large-v1} \cite{rerank2024mxbai}, and no reranking as a baseline.

\subsection{Stance Generation}

The final step generates a score and a reason for the retrieved evidences. To ensure privacy and control, we use open-weight Qwen3 models \cite{yang2025qwen3technicalreport} and compare different zero- and few-shot prompting strategies (see Appendix \ref{appendix:a}).

\section{Evaluation}
\subsection{Dataset}

We evaluate our system on a proprietary dataset constructed in collaboration with InfluenceMap. The data mirrors the structure proposed in \citet{NEURIPS2023_7ccaa4f9}, where each instance aligns manually selected textual evidence with a climate policy query and classifies its stance (see Appendix \ref{sec:data_structure}).

\subsection{Metrics}

\textbf{Normalized Longest Common Subsequence (nLCS)} quantifies textual overlap between sequences, producing a score between 0 (no match) and 1 (perfect match). The LCS length is computed using Algorithm~\ref{alg:lcs}, and the result is normalized differently depending on the task.

\textbf{Recall and MRR} are metrics to evaluate the retriever and reranker components. \cite{weaviateRetrievalMetrics2023}

\textbf{Exact Match Accuracy} considers the stance classifier predictions a hit if:
\begin{equation}
\text{EM}(y, \hat{y}) = \mathbbm{1}_{\left[\hat{y} = y\right]}
\end{equation}

\textbf{Hit Rate with Tolerance} is a relaxed criterion that considers the prediction a hit with a tolerance $\tau = 1$ conditioned on polarity alignment:
\begin{equation}
\text{HRT}(y, \hat{y}) = 
\begin{cases}
1 & \parbox[t]{.5\linewidth}{if $|y - \hat{y}| \leq \tau$ and $\text{sign}(y) = \text{sign}(\hat{y})$} \\
1 & \text{if } y = \hat{y} = 0 \\
0 & otherwise
\end{cases}
\label{eq:hrt}
\end{equation}

\textbf{Oracle-Based Diagnostics} assess evidence extraction pipelines through correlations between prediction errors and individual RAG component outputs \cite{zhao2024seerselfalignedevidenceextraction}. Faithfulness\footnote{We use AlignScore-large, similar to \citet{zhao2024seerselfalignedevidenceextraction}} \cite{zha-etal-2023-alignscore} and conciseness\footnote{We use the bge-m3 model with the same formula in \citet{zhao2024seerselfalignedevidenceextraction}} scores, $s_f$ and $s_c$ respectively,  compare the alignment of the retrieved snippet $e$ with the gold snippet $G$. The helpfulness\footnote{We apply changes as shown in equation \ref{eq:help} to the original formula in \citet{zhao2024seerselfalignedevidenceextraction}} $s_h$ measures the confidence of the model in generating the correct stance $a$ using  $G$ vs. $e$:

\begin{equation}
s_h = \sigma \left( \log \frac{f(y \mid q, G)}{f(y \mid q, e)} \right)
\label{eq:help}
\end{equation}

\noindent
where $\sigma$ is the sigmoid function, and $f(y \mid q, x)$ is the model-assigned probability of the stance answer $y$ given the query $q$ and $x=G$ or $x=e$.

\subsection{Component Evaluation}

\textbf{Parsing and Chunking} are evaluated using $nLCS$. In parsing evaluation $P_{nLCS}$, normalization is based on the length of $G$ to assess whether the parser can recover the reference content anywhere in the output text $P$, tolerating additional surrounding text. 

In chunking evaluation $C_{nLCS}$, we compare the chunks produced $\{C_i\}$ from $P$ against $G$, for files that achieve a parsing score of higher than threshold $\sigma$. Normalization uses the length of the longer sequence to penalize both omissions and excessive context. 

This distinction allows parsing to reward inclusion, while ensuring that chunking emphasizes precision.

\textbf{Retrieval and Reranking} are evaluated using Recall and MRR metrics on files achieving a higher parsing and chunking score than threshold $\sigma$. $nLCS$ between retrieved $e$ and $G$ is used to determine relevant retrievals.

\textbf{Stance Generation} uses Exact Match Accuracy and Hit Rate with Tolerance comparing generated label $\hat{y}$ from $G$ with human annotation $y$.

\subsection{Pipeline Evaluation}
\label{sec:pe}
We compare four retrieval strategies to evaluate the pipeline on the final stance generation task:

\begin{enumerate}
    \item \textbf{First Retrieved (FR)}: Using the top result from semantic retrieval.
    \item \textbf{All Retrieved (AR)}: Concatenating all top-k chunks.
    \item \textbf{Best Match (BM)}: Using the chunk with highest nLCS overlap with $G$. Useful to showcase the impact of rank. 
    \item \textbf{Ground Truth (GT)}: Serves as an upper bound.
\end{enumerate}

We implement Oracle diagnostic metrics to understand: the degree of factual consistency ($s_f$), the certainty of the model during stance generation ($s_h$), and the degree of information gap or overload ($s_c$).

\section{Results}
Our goal is to return the most relevant evidence first and classify it correctly. We evaluated both the complete pipeline and individual components.

\subsection{Parsing and Chunking}

In Table \ref{tab:parser_chunking_detail}, we report the parsing and chunking results across multilingual documents. 

\begin{table}[h]
\centering
\small
\renewcommand{\arraystretch}{1.1}
\resizebox{\columnwidth}{!}{%
\setlength{\tabcolsep}{1pt}
\begin{tabular}{llcccccc}
\toprule
\textbf{Parser} & \textbf{Lang.} & Chunks–L & Chunks–S & $P_{nLCS}$ & $C_{nLCS}$–L & $C_{nLCS}$–S \\
\midrule
\multirow{3}{*}{Docling} 
  & All    & 22.3 & 46.1 & \textbf{0.673} & \textbf{0.404} & \textbf{0.357} \\
  & EN     & 25.1 & 51.7 & 0.695\textsuperscript{*} & 0.459\textsuperscript{*} & 0.401\textsuperscript{*} \\
  & Non-EN & 11.3 & 23.6 & 0.587 & 0.188 & 0.181 \\
\midrule
\multirow{3}{*}{PyMuPDF} 
  & All    & 54.5 & 126.7 & 0.669 & 0.325 & 0.329 \\
  & EN     & 60.3 & 145.7 & 0.642 & 0.357 & 0.359 \\
  & Non-EN & 31.4 & 51.4 & 0.774\textsuperscript{+} & 0.200\textsuperscript{+} & 0.209\textsuperscript{+} \\
\bottomrule
\end{tabular}
}
\caption{Parsing fidelity ($P_{nLCS}$), chunk compactness ($C_{nLCS}$), and average number of chunks using layout (–L) and semantic (–S) chunking strategies across language groups include \textbf{All languages}, English-only\textsuperscript{*}, and Non-English\textsuperscript{+}.}
\label{tab:parser_chunking_detail}
\end{table}

Docling outperforms PyMuPDF in parsing fidelity for both English and the overall dataset, confirming its strength in layout-sensitive documents. However, PyMuPDF demonstrates superior parsing performance on non-English texts. When comparing chunking strategies, layout-based chunking consistently outperforms semantic chunking when applied to Docling outputs. In contrast, it performs worst when used on PyMuPDF-parsed text, due to its layout-agnostic nature. The highest chunking fidelity for the full set is achieved using Docling combined with layout chunking.

\subsection{Retrieval Performance}
We evaluate the retriever across the different parsing strategies and chunking methods. To isolate retriever quality, we include only files where $P_{nLCS}$ exceeds $\sigma=0.5$, as retrieved chunks are considered \textit{relevant} if their $nLCS$ with $G$ exceeds this threshold. Table \ref{tab:retrieval_results} summarizes the results.

\begin{table*}[ht]
\centering
\small
\renewcommand{\arraystretch}{1.2}
\setlength{\tabcolsep}{4pt}
\begin{tabular}{ll|ccc|ccc|ccc|ccc}
\toprule
\multirow{2}{*}{\textbf{Embedding}} & \multirow{2}{*}{\textbf{Lang.}} 
& \multicolumn{3}{c|}{\textbf{Docling – Layout}} 
& \multicolumn{3}{c|}{\textbf{Docling – Semantic}} 
& \multicolumn{3}{c|}{\textbf{PyMuPDF – Layout}} 
& \multicolumn{3}{c}{\textbf{PyMuPDF – Semantic}} \\
\cmidrule(lr){3-5} \cmidrule(lr){6-8} \cmidrule(lr){9-11} \cmidrule(lr){12-14}
& & $nLCS$ & Recall & $C_{nLCS}$ 
& $nLCS$ & Recall & $C_{nLCS}$ 
& $nLCS$ & Recall & $C_{nLCS}$ 
& $nLCS$ & Recall & $C_{nLCS}$ \\
\midrule
\multirow{3}{*}{bge-m3}
& All        & 0.580 & 0.543 & 0.378 & 0.643 & 0.645 & 0.300 & 0.662 & 0.666 & 0.241 & \textbf{0.468} & \textbf{0.434} & \textbf{0.301} \\
& English    & 0.542 & 0.495 & 0.422 & 0.625 & 0.622 & 0.326 & 0.627 & 0.629 & 0.243 & 0.403\textsuperscript{*} & 0.356\textsuperscript{*} & 0.320\textsuperscript{*} \\
& Non-EN     & 0.770\textsuperscript{+} & 0.776 & 0.164\textsuperscript{+} & 0.729 & 0.759 & 0.176 & 0.836 & 0.845 & 0.232\textsuperscript{+} & 0.791 & 0.819 & 0.211 \\
\midrule
\multirow{3}{*}{nomic}
& All        & \textbf{0.592} & \textbf{0.564} & \textbf{0.389} & \textbf{0.724} & \textbf{0.764} & \textbf{0.312} & 0.696 & 0.730 & 0.245 & 0.435 & 0.393 & 0.283 \\
& English    & 0.557\textsuperscript{*} & 0.517\textsuperscript{*} & 0.436\textsuperscript{*} & 0.724\textsuperscript{*} & 0.760\textsuperscript{*} & 0.339\textsuperscript{*} & 0.672 & 0.703 & 0.252 & 0.376 & 0.322 & 0.302 \\
& Non-EN     & 0.769 & 0.793\textsuperscript{+} & 0.154 & 0.723 & 0.784 & 0.180 & 0.814 & 0.862 & 0.209 & 0.724 & 0.741 & 0.190 \\
\midrule
\multirow{3}{*}{qwen}
& All        & 0.547 & 0.488 & 0.354 & 0.719 & 0.757 & 0.279 & \textbf{0.716} & \textbf{0.738} & \textbf{0.250} & 0.434 & 0.381 & 0.263 \\
& English    & 0.505 & 0.426 & 0.397 & 0.697 & 0.723 & 0.298 & 0.685\textsuperscript{*} & 0.706\textsuperscript{*} & 0.257\textsuperscript{*} & 0.337 & 0.271 & 0.270 \\
& Non-EN     & 0.757 & 0.793\textsuperscript{+} & 0.144 & 0.831\textsuperscript{+} & 0.922\textsuperscript{+} & 0.187\textsuperscript{+} & 0.866\textsuperscript{+} & 0.897\textsuperscript{+} & 0.218 & 0.908\textsuperscript{+} & 0.922\textsuperscript{+} & 0.227\textsuperscript{+} \\
\bottomrule
\end{tabular}
\caption{Evaluation of retrieval quality across different embedding models (bge-m3, nomic, qwen), parsing strategies (Docling vs PyMuPDF), and chunking methods (layout vs semantic). Metrics include $nLCS$ with gold evidence, recall@5, and $C_{nLCS}$. Language groups include \textbf{All languages}, English-only\textsuperscript{*}, and Non-English\textsuperscript{+}.}
\label{tab:retrieval_results}
\end{table*}

Across the combined and English language sets, Nomic consistently delivers the best retrieval performance when paired with Docling parsing, particularly under semantic chunking. It achieves top scores in both $nLCS$ (0.724) and Recall (0.764) while maintaining a relatively high $C_{nLCS}$ (0.312), indicating focused, relevant chunk retrieval. Qwen surpasses Nomic on the non-English subset, especially with semantic chunking, where it reaches the highest Recall (0.922). Qwen’s multilingual edge reflects Nomic’s English-centric training. However, its high recall coincides with lower $C_{nLCS}$, indicating retrieval of broader, less precise evidence.

Non-English retrieval consistently outperforms due to lower chunk counts (Table\ref{tab:parser_chunking_detail}), which reduces the vector search space. This structural bias inflates recall by making it easier to hit a relevant chunk while simultaneously lowering $C_{nLCS}$ due to increased dilution.

PyMuPDF with layout chunking performs best with Qwen on non-English texts, where coarser segmentation enhances recall but increases $C_{nLCS}$. BGE performs consistently but does not lead on any metric.

Overall, Docling provides the best trade-off between $C_{nLCS}$ and Recall, with layout chunking offering superior chunk precision, and semantic chunking optimizing retriever effectiveness. Nomic stands out as the most robust embedding model across configurations, while Qwen excels in recall-heavy, high chunk dilution scenarios, especially for non-English content.

\subsection{Reranker Performance}

To assess reranker performance in isolation, we conduct all experiments on the vector spaces generated from Docling-parsed, layout- or semantic-based chunks embedded using the Nomic model, the best-performing retriever configurations. This setup ensures a consistent and high-quality evidence pool, allowing us to isolate and compare the effectiveness of different reranking strategies. Table \ref{tab:reranker_mrr} summarizes the results.

\begin{table}[h]
\centering
\small
\renewcommand{\arraystretch}{1.1}
\resizebox{\columnwidth}{!}{%
\setlength{\tabcolsep}{4pt}
\begin{tabular}{llcccccc}
\toprule
\textbf{Reranker} & \textbf{Language} & \textbf{Layout} & \textbf{Semantic} \\
\midrule
No Reranker & All         & 0.359 & 0.490 \\
            & English     & 0.296 & 0.491 \\
            & Non-English & 0.603\textsuperscript{+} & 0.485 \\
\midrule
BGE         & All         & \textbf{0.374} & \textbf{0.531} \\
            & English     & 0.321\textsuperscript{*} & 0.528\textsuperscript{*} \\
            & Non-English & 0.577 & 0.544\textsuperscript{+} \\
\midrule
MXBAI       & All         & 0.359 & 0.490 \\
            & English     & 0.296 & 0.491 \\
            & Non-English & 0.603\textsuperscript{+} & 0.485 \\
\bottomrule
\end{tabular}
}
\caption{Mean Reciprocal Rank (MRR) for different reranking strategies applied to Docling-parsed, Nomic-embedded chunks across chunking methods and language groups: \textbf{All languages}, English-only\textsuperscript{*}, and Non-English\textsuperscript{+}.}
\label{tab:reranker_mrr}
\end{table}

Across both chunking strategies, BGE consistently outperforms the no-reranker baseline, particularly in the semantic setup, where it achieves the highest MRR on the full (0.531) and English sets (0.528). In contrast, MXBAI provides no improvement, matching the baseline in all cases. Notably, non-English layout chunks achieve the highest MRR (0.603) even without reranking, reflecting the inherent ease of retrieval in these documents due to longer, fewer chunks. This limits the reranker's impact and aligns with prior findings. Overall, semantic chunking benefits most from reranking, and BGE emerges as the most effective strategy when higher precision is required.

\subsection{Stance Generation Performance}
\label{llm:stance_generation}
We assess Qwen3’s ability to generate accurate stance predictions when provided with the gold evidence snippets. This isolates generation quality from retrieval noise, establishing an upper bound for stance prediction performance.

\begin{figure}
    \includegraphics[scale=0.2]{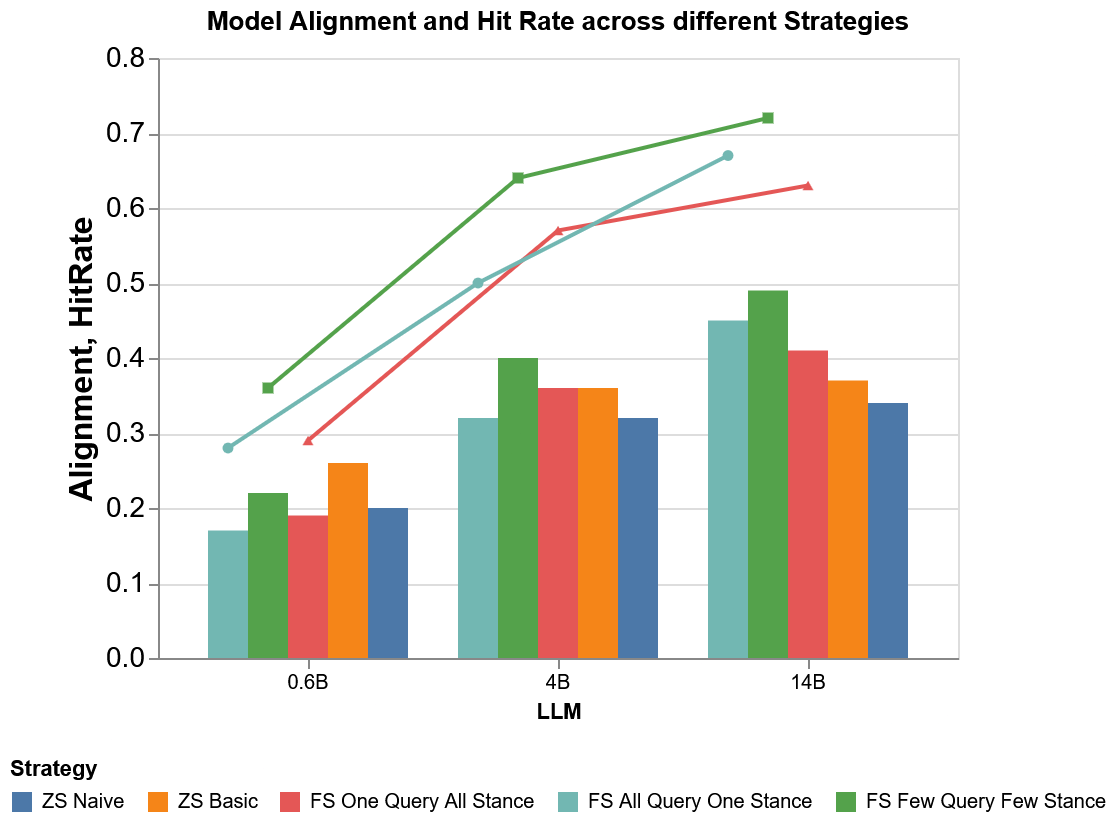}
    \caption{Model Alignment and Hit Rates}
    \label{alignment}
\end{figure}

Figure \ref{alignment} shows that alignment improves steadily with model size across all strategies, indicating that larger models are better at inferring correct stances from ground truth evidence. Among few-shot (FS) strategies, "Few Query Few Stance" achieves the highest alignment at both the 4B and 14B scales, suggesting that since some queries are similar to each other, exposing the model to a few representative queries, even when focused on a single stance, provides strong grounding. 

Zero-shot (ZS) strategies lag behind FS across all model sizes. Interestingly, "ZS" strategies perform competitively at 0.6B, but its advantage diminishes with scale, reaffirming that smaller models benefit from shorter contexts, while larger models thrive with richer few-shot demonstrations.

For "FS" strategies, we also provide Hit Rate, which tells an analyst whether the evidence is expected to support or oppose the query. We observe that Hit Rates can reach 0.70, which once again highlights the usefulness of LLMs for this task.

See Appendix \ref{appendix:llm_eval} for more details on LLM evaluation.
\subsection{Pipeline Performance}
We evaluate the full pipeline using the best-performing configuration: Docling parsing, semantic chunking, Nomic embeddings, BGE reranking, and Qwen3:4B "FS Few Query Few Stance" prompting. We compare the four evidence selection strategies that vary in relevance, ranking position, and chunk quality (Section \ref{sec:pe}).

\begin{table}[h]
\centering
\small
\renewcommand{\arraystretch}{1.2}
\begin{tabular}{lcccc}
\toprule
\textbf{Strategy} & \textbf{Hit Rate} & \textbf{Exact Match} \\
\midrule
Ground Truth (GT)   & 0.673 & 0.345 \\
First Retrieved (FR)     & 0.636 & 0.309 \\
Best Match (BM)         & 0.600 & 0.309 \\
All Retrieved (AR)       & 0.655 & 0.273 \\
\bottomrule
\end{tabular}
\caption{Stance generation accuracy (Hit Rate and Exact Match) across retrieval strategies using Docling-Semantic chunks, Nomic embeddings, and BGE reranking.}
\label{tab:stance_results}
\end{table}

The results in Table \ref{tab:stance_results} show that GT yields the highest stance accuracy, confirming the upper bound of LLM performance. Among retrieval strategies, AR performs best, nearly matching the hit rate of Ground Truth, despite its lower exact match. This suggests that aggregating multiple chunks increases the chance of including relevant content, but at the cost of precision. FR consistently outperforms BM in both metrics, highlighting the strength of BGE reranking in surfacing useful evidence early. However, the narrow gap across all methods indicates that chunk quality, rather than just retrieval rank, plays a critical role in final stance reliability.

To better understand why stance prediction succeeds or fails under each retrieval strategy, we complement accuracy metrics with oracle-based diagnostics in Table \ref{tab:oracle_scores}.

\begin{table}[ht]
\centering
\small
\renewcommand{\arraystretch}{1.2}
\begin{tabular}{lccc}
\toprule
\textbf{Metric} & \textbf{BM} & \textbf{FR} & \textbf{AR} \\
\midrule
Faithfulness ($s_f$) & 0.753 & 0.420 & 0.830 \\
Helpfulness ($s_h$)  & 0.561 & 0.582 & 0.415 \\
Conciseness ($s_c$)  & 0.855 & 0.737 & 0.760 \\
\bottomrule
\end{tabular}
\caption{Average oracle diagnostic scores for the retrieval strategies: Best Match (BM), First Retrieved (FR), and All Retrieved (AR).}
\label{tab:oracle_scores}
\end{table}

Despite its lower faithfulness score ($s_f = 0.420$), the FR strategy achieves the highest stance generation accuracy. This aligns with its top helpfulness score ($s_h = 0.582$), suggesting that early-ranked chunks, though not always the most faithful to ground truth, can still effectively guide the model to the correct stance. In contrast, AR achieves the highest faithfulness ($s_f = 0.830$) but performs worst in Exact Match, likely due to lower helpfulness ($s_h = 0.415$) and moderate conciseness, indicating that excessive context may dilute the model’s focus. BM balances strong faithfulness ($s_f = 0.753$) and top conciseness ($s_c = 0.855$), but its lower hit rate suggests that even aligned and compact evidence may fall short if not supported by helpfulness. These results reveal a critical trade-off: while faithfulness correlates with gold overlap, it does not always translate to better stance prediction. Helpfulness, particularly for early-ranked evidence, appears more decisive for final performance.

\paragraph{When Retrieval Outperforms Ground Truth.}
To understand when retrieved evidence outperforms gold snippets, we isolate cases where stance prediction fails using the GT snippet but succeeds with at least one of the retrieval strategies, as measured by the Exact Match metric. We compute average oracle scores for these outperforming cases by retrieval strategy (see Table~\ref{tab:oracle_outperforming} in the Appendix).

The results challenge the assumption that gold-labeled evidence is always optimal for automated LLM stance generation. Notably, FR accounts for nearly half of all outperforming cases and achieves the highest helpfulness score, suggesting that early-ranked evidence, despite lower faithfulness, can guide the LLM toward correct stance generation. AR achieves the highest faithfulness, reflecting its ability to recover relevant information through context aggregation, though it may introduce information overload. BM is the most concise and moderately faithful to GT, thus limiting its ability to outperform it. This highlights that success depends not just on gold alignment, but on how evidence supports reasoning, by providing the context LLMs actually need.

\section{Conclusion}

This paper presents a multilingual RAG system for automated corporate climate policy evidence extraction and scoring. Through careful assessment of the components, we demonstrate that multilingual, layout-aware parsing combined with semantic chunking and reranked dense retrieval yields high-quality evidence for LLM-based stance inference.

Our experiments highlight the strengths and trade-offs of different retrieval strategies, showing that faithfulness alone does not guarantee optimal performance. Notably, retrieved evidence, especially when well-ranked or aggregated, can sometimes outperform human-annotated gold snippets, suggesting that LLMs benefit from context that extends beyond manually defined spans. These findings underscore the importance of adaptive retrieval pipelines that prioritize not just relevance, but evidence helpfulness and conciseness to support accurate and explainable LLM reasoning.


Future work should focus on improving ranking algorithms for fine-grained chunks and developing better evaluation metrics that account for the inherent challenges of precise evidence extraction from complex corporate documents. 


\bibliographystyle{acl_natbib}
\bibliography{emnlp2023-latex/anthology}
\appendix
\section*{Limitations}


This work has several limitations. First, the evaluation relies on a fixed set of policy queries and gold annotations, which may not capture the full variability of corporate discourse or user intent, limiting its generalizability. Second, the stance prediction model is sensitive to prompting and may overfit to the specific formulation used in few-shot examples. Third, while our oracle metrics provide insight into evidence quality, they are themselves approximations and may not fully reflect nuanced LLM reasoning failures.
Finally, the system currently lacks explicit mechanisms to detect annotation errors or leverage multi-document context, both of which could improve robustness in real-world use cases.

\section*{Ethics Statement}

This research supports transparency in corporate climate policy engagement, which serves the public interest. The automated system is designed to augment rather than replace human analyst judgment. All data processing respects privacy considerations, and the system is intended to promote accountability in climate policy discourse.

\section{Prompting Strategies}
\label{appendix:a}
Naive Prompt:
\begin{lstlisting}
system_naive: |-
  You are an expert climate analyst specializing in assessing companies' engagement stances on climate policies. You can only give a score between -2 and 2 where the scores mean as follows :
  -2: opposing
  -1: not supporting
  0: no position/mixed position
  1: supporting
  2: strongly supporting

  **Important**:
  - You must always use the report_stance tool provided to analyze the evidence and determine a company's position on the provided specific climate policy query.
  - Do not respond without using the tool.
  - You can only give a score between -2 and 2.
\end{lstlisting}

Basic Prompt:
\begin{lstlisting}
system_basic: |-
  You are an expert climate analyst specializing in assessing companies' engagement stances on climate policies. You can only give a score between -2 and 2 where the scores mean as follows :
  -2: Position contradicts IPCC analysis OR Opposes policy
  -1: Position appears misaligned with IPCC analysis OR Unsupportive of policy and/or communicates support for the policy but with major caveats and/or conditions that would weaken the strength of the proposal
   0: Unclear if position is aligned with IPCC guidance OR Unclear if position is supportive of policy
  +1: Broad alignment with IPCC analysis OR General or high-level support for the policy
  +2: Detailed position that is aligned with IPCC analysis OR Strong Support for the policy and/or advocacy that would strengthen the policy further

  **Important**:
  - You must always use the report_stance tool provided to analyze the evidence and determine a company's position on the provided specific climate policy query.
  - Do not respond without using the tool.
  - You can only give a score between -2 and 2.    
\end{lstlisting}
One query with all stances
\begin{lstlisting}
system_one_query_all_stance: |-
  You are an expert climate analyst specializing in assessing companies' engagement stances on climate policies. You can only give a score between -2 and 2 where the scores mean as follows :
  -2: Position contradicts IPCC analysis OR Opposes policy
  -1: Position appears misaligned with IPCC analysis OR Unsupportive of policy and/or communicates support for the policy but with major caveats and/or conditions that would weaken the strength of the proposal
   0: Unclear if position is aligned with IPCC guidance OR Unclear if position is supportive of policy
  +1: Broad alignment with IPCC analysis OR General or high-level support for the policy
  +2: Detailed position that is aligned with IPCC analysis OR Strong Support for the policy and/or advocacy that would strengthen the policy further

  **Important**:
  - You must always use the report_stance tool provided to analyze the evidence and determine a company's position on the provided specific climate policy query.
  - Do not respond without using the tool.
  - You can only give a score between -2 and 2.

  Below are examples of how to rate stance on a query using given context.
  ---
  Question: Is the organization supporting an IPCC-aligned transition of the economy away from carbon-emitting technologies, including supporting relevant policy and legislative measures to enable this transition?

  Context: Combustion of fossil fuels gives rise to almost 90 percent of all carbon dioxide emissions. An established time frame for the phasing out of fossil fuels in all sectors within the EU would contribute to increased clarity and more predictable rules of the game for business and society's other actors. Russia's invasion of Ukraine has also underlined the vulnerability of continued European dependence on fossil fuels. Decisions to phase out fossil fuels within the EU benefit energy security, public health and the development of a sustainable and viable business that is competitive even in the long term. [...] Phase out all use of fossil fuels within the EU. Sweden should work for a complete phasing out of fossil fuels within the EU through decisions which mean that the use of coal ceases around 2030, natural gas is phased out in the mid-2030s (not biogas) and oil by 2040. All fossil fuel subsidies within the EU should also quickly cease, at the same time that vulnerable groups in society who in the short term suffer from increased costs as a result of climate measures are compensated.

  Stance: +2

  ---
  Question: Is the organization supporting an IPCC-aligned transition of the economy away from carbon-emitting technologies, including supporting relevant policy and legislative measures to enable this transition?

  Context: Naturally, we complied with the order, but we believe that we, as a society, must phase out the use of gas, oil, and coal as soon as possible, and with the close down of the heat and power plant, we’re well on track to becoming the first major energy company to completely transform its energy production from fossil fuels to renewable energy.

  Stance: +1

  ---
  Question: Is the organization supporting an IPCC-aligned transition of the economy away from carbon-emitting technologies, including supporting relevant policy and legislative measures to enable this transition?

  Context: Transcript: And deliver that by building on our industrial heritage. We are so lucky here in the UK, we’ve got a fantastic oil and gas sector, we’ve got the second largest wind farm capability in the world, 200,000 brilliant people working up and down the country. We need to unleash that. [...] Let’s be clear. What we’re not talking about is expansion of oil and gas. The North Sea has powered out economy for the last 50 years, but it’s very clear this is a declining basin. So oil and gas volumes that we produce here in the UK are going to decline. Today, our country, we need 75% of our energy comes from oil and gas, and we produce about half of that. There isn’t a scenario where we’re going to grow our oil and gas production, but what we call for is a really managed transition. While we still use oil and gas, surely it makes sense that we use our own homegrown energy. And if we do that, we support the people, those important supply chain companies, that actually not only will help us with our oil and gas, but will pivot to develop our world class capability in carbon storage, in wind, particularly floating wind, and also hydrogen. 

  Stance: 0

  ---
  Question: Is the organization supporting an IPCC-aligned transition of the economy away from carbon-emitting technologies, including supporting relevant policy and legislative measures to enable this transition?

  Context: "Oil-to-Specialties" refers to the gradual increase in the share of refining crude oil directly into specialty products such as base oils for lubricants, white oils, sulfur, asphalt, and petroleum coke [...] Driven by the carbon peak and carbon neutrality goals, specialty oil products will continue to play a significant role in society and the economy. From a market perspective, these products have broad applications, spanning traditional industries and strategic emerging sectors. Not only will future demand for these specialty oils maintain steady growth, but they are also crucial for preserving the completeness of China's industrial system and supporting the autonomous development of the national economy. Additionally, many specialty oil products are highly profitable, and due to the diverse downstream industries involved, their overall demand is resilient and risk-resistant. This makes them a potential long-term profit driver and backup resource for petrochemical enterprises. 

  Stance: -1

  ---
  Question: Is the organization supporting an IPCC-aligned transition of the economy away from carbon-emitting technologies, including supporting relevant policy and legislative measures to enable this transition?

  Context: The U.S. upstream industry is safely and sustainably delivering the energy the world needs, while also leading in the reduction of Green House Gas (GHG) emissions. We believe that the U.S. federal government and U.S. allies should strongly support continued development of top-quality U.S. oil and gas resources to help meet the current and future energy demands of the world.  

  Stance: -2

  ---
\end{lstlisting}
One Query All Stance
\begin{lstlisting}
system_all_query_one_stance: |-
  You are an expert climate analyst specializing in assessing companies' engagement stances on climate policies. You can only give a score between -2 and 2 where the scores mean as follows :
  -2: Position contradicts IPCC analysis OR Opposes policy
  -1: Position appears misaligned with IPCC analysis OR Unsupportive of policy and/or communicates support for the policy but with major caveats and/or conditions that would weaken the strength of the proposal
  0: Unclear if position is aligned with IPCC guidance OR Unclear if position is supportive of policy
  +1: Broad alignment with IPCC analysis OR General or high-level support for the policy
  +2: Detailed position that is aligned with IPCC analysis OR Strong Support for the policy and/or advocacy that would strengthen the policy further

  **Important**:
  - You must always use the report_stance tool provided to analyze the evidence and determine a company's position on the provided specific climate policy query.
  - Do not respond without using the tool.
  - You can only give a score between -2 and 2.
  
  ---
  Question: Is the organization transparent and clear about its position on climate change science?

  Context: 216 million people could be forced to migrate within their own countries by 2050 because of climate change. [...] Switching to a clean economy could raise $2.8 trillion through carbon price revenues and the redirection of fossil fuel subsidies to public investments [...] To avert the worst impacts of climate change, global temperature increase must be limited to 1.5C above pre-industrial levels. Global temperatures are already 1.1C warmer.  [...] Indonesia and the U.S. are the countries with the highest shares of climate change deniers.

  Stance: +2

  ---
  Question: Is the organization supporting the science-based response to climate change as set out by the IPCC?

  Context: As the world’s largest logistics service provider we want to reduce all logistics-related emissions to net-zero by 2050. Thus, we contribute to achieving the Paris Climate Protection Agreement of reducing global warming to below 2C, and we support the United Nation’s Agenda 2030.  

  Stance: +1
  ---

  Question: Is the organization supporting the UN FCCC process on climate change?

  Context: The Alliance is looking to governments to accelerate the transition before COP26 and beyond and calls on world leaders to:**Publish ambitious and 1.5C-aligned Nationally Determined Contributions that halve emissions by 2030**Commit to net-zero by 2050, underpinned by robust policy roadmaps and interim targets**Ensure that developed countries meet and exceed their $100B commitment to support developing countries mitigate and adapt to climate change, and ensure the major development finance institutions also commit to science-based guidelines across their lending portfolios  

  Stance: +2
  
  ---
  Question: To what extent does the organization express the need for regulatory intervention to resolve the climate crisis?

  Context: The scale of climate issues is such that global progress requires significant cooperation between the private and public sectors. Engagement with policymakers and regulators enables Dell to learn how it can best support global efforts, to communicate where external support can make Dell’s work on climate more effective, and to work with governments to develop and implement climate solutions. To that extent, Dell has identified a number of areas where policy initiatives can help Dell to leverage our capabilities and portfolio for accelerated global progress on climate. Dell will work with public organizations on these topics and will be a public advocate to help communicate the potential for policy to drive progress on climate issues.  

  Stance: 0

  ---
  Question: Is the organization transparent about its positions on climate change legislation/policy and its activities to influence it?

  Context: On what policy, law, or regulation that may impact the climate has your organization been engaging directly with policy makers in the reporting year? [Specify the policy, law, or regulation on which your organization is engaging with policy makers] F-Gas Regulation EU 517/2014 and UK equivalent [...] [Specify the policy, law, or regulation on which your organization is engaging with policy makers] Regulation for the deployment of alternative fuels infrastructure (AFIR) [...] [Specify the policy, law, or regulation on which your organization is engaging with policy makers] Electric Vehicle Infrastructure Strategy [...] [Specify the policy, law, or regulation on which your organization is engaging with policy makers] SI 2021/1242: Road Vehicle Carbon Dioxide Emission Performance Standards (Cars and Vans) (Miscellaneous Amendments) Regulations 2021 [...] [Specify the policy, law, or regulation on which your organization is engaging with policy makers] CO2 Fleet Regulation (EU 2019/631) finally amended by regulation EU 2023/851 (trialogue outcome) [...] [Specify the policy, law, or regulation on which your organization is engaging with policy makers] UN ECE Life cycle assessment [...] [Specify the policy, law, or regulation on which your organization is engaging with policy makers] Euro 7 (replacing 715/2007) [...]

  Stance: -2

  ---
  Question: Is the organization supporting policy and legislative measures to address climate change: carbon tax?

  Context: We agree with the principle that an economy wide price on carbon is the best way to use the power of the market to achieve carbon reduction goals, in a simple, coherent and efficient manner. Markets will also spur innovation and create and preserve quality jobs in a growing low-carbon economy. 

  Stance: +1

  ---
  Question: Is the organization supporting policy and legislative measures to address climate change: emissions trading?

  Context: The company also actively monitors the legislative and regulatory processes to help inform its investment decisions. For example, legislation to address climate change by reducing greenhouse gas emissions and establishing a price on carbon could create increases in energy costs and price volatility. There are existing efforts to address GHG emissions at the national and regional levels. Several of the company's facilities in the European Union (EU) are regulated under the EU Emissions Trading Scheme. China has begun pilot programs for carbon taxes and trading of GHG emissions in selected areas. In the EU, U.S. and Japan, policy efforts to reduce the GHG emissions associated with gases used in refrigeration and air conditioning create market opportunities for lower GHG solutions. 

  Stance: 0

  ---
  Question: Is the organization supporting policy and legislative measures to address climate change: energy efficiency policy, standards, and targets?

  Context: Mandatory targets are effective as witnessed by the progress towards achieving the 20-20-20 targets. The EU is on track to meet its renewables objectives and its decarbonisation objectives but is lagging behind on energy efficiency, the sole target of the 2009 Climate Action and Renewable Energy (CARE) Package that is not binding. 

  Stance: +1

  ---
  Question: Is the organization supporting policy and legislative measures to address climate change: Renewable energy legislation, targets, subsidies, and other policy?

  Context: A group of big technology, industry and power companies have called on the European Union to set a target for renewables of at least 35 percent when EU energy ministers meet next week. The energy-hungry firms, including Amazon, Facebook, Google , IKEA, Microsoft, Philips and Unilever, say an ambitious target would encourage their investment in multi-year wind and solar power supply contracts, known as Power Purchase Agreements (PPAs). In a letter, the 50 big firms called on EU energy ministers to lift all regulatory barriers to PPAs, to which firms are increasingly turning to source electricity needed for energy-intensive data centers or to run heavy machinery. 

  Stance: +2
  
  ---
  Question: Is the organization supporting an IPCC-aligned transition of the economy away from carbon-emitting technologies, including supporting relevant policy and legislative measures to enable this transition?

  Context: [16. Specific lobbying issues:] Lobbied in support of HR 4468, the Choice in Automobile Retail Sales Act, which limits the Environmental Protection Agency's ability to mandate the use of a specific technology and results in limited availability of new motor vehicles. 
 
  Stance: -2

  ---
  Question: Is the organization supporting policy and legislative measures to address climate change: standards, targets, and other regulatory measures directly targeting Greenhouse Gas emissions?

  Context: Peabody supports adoption of the Affordable Clean Energy (ACE) rule as a replacement for the Clean Power Plan as a significant step in ensuring reliable, resilient and affordable electricity across the U.S. ACE strengthens regulatory and investment certainty using a sensible and legally defensible "within the fenceline" approach to improve efficiencies of existing power plants. The proposed rule offers technically feasible and appropriate measures with the potential to deliver cost-effective, achievable and practical solutions for reducing emissions while minimizing disruptions to electricity generation. Furthermore, ACE offers individual states greater flexibility in the development and timing of state implementation plans, avoiding a "one-size-fits-all" approach to managing distinct and diverse needs. Peabody has long advocated for technology solutions to meet the three-part goal of energy security, economic progress and environmental solutions and we believe ACE satisfies these objectives. 

  Stance: -2

  ---
  Question: Is the organization transparent about its involvement with industry associations that are influencing climate policy, including the extent to which it is aligned with these groups on climate?

  Context: Evergy employees serve on multiple EEI committees and in leadership positions on these committees. EEI is the association that represents all U.S. investor-owned electric companies. EEI provides public policy leadership, strategic business intelligence, and essential conferences and forums. EEI’s member companies are leading a clean energy transformation. [...] One example of a policy position Evergy supports and has been instrumental in moving forward: In December 2021, EEI launched the National Electric Highway Coalition (NEHC), a collaboration among electric companies, including Evergy, that are committed to providing EV fast charging stations allowing the public to drive EVs with confidence along major U.S. travel corridors by the end of 2023. The NEHC is the largest such alliance of electric companies that have organized around the goal of deploying EV fast charging infrastructure to support the growing number of EVs and ensure that the transition to EVs is seamless for drivers. 

  Stance: -1

  ---
  Question: Is the organization supporting policy and legislative measures to enhance and protect ecosystems and land where carbon is being stored?

  Context: Supporting a diverse portfolio of natural and technological carbon removal projects is essential to maximize near-term climate impact while supporting necessary carbon removal solutions for the future. Nature-based projects are available now, begin sequestering carbon within the first years of implementation, and can provide positive local impacts like supporting community resilience or increasing the ecological health of a region. Emerging technologies like direct air capture, which have a high global climate mitigation potential and can offer durable carbon storage, will be a critical complement to nature-based removals in enabling a zero carbon future. 

  Stance: 0

  ---
\end{lstlisting}

Few Query Few Stance
\begin{lstlisting}
system_few_query_few_stance: |-
  You are an expert climate analyst specializing in assessing companies' engagement stances on climate policies. You can only give a score between -2 and 2 where the scores mean as follows :
  -2: Position contradicts IPCC analysis OR Opposes policy
  -1: Position appears misaligned with IPCC analysis OR Unsupportive of policy and/or communicates support for the policy but with major caveats and/or conditions that would weaken the strength of the proposal
  0: Unclear if position is aligned with IPCC guidance OR Unclear if position is supportive of policy
  +1: Broad alignment with IPCC analysis OR General or high-level support for the policy
  +2: Detailed position that is aligned with IPCC analysis OR Strong Support for the policy and/or advocacy that would strengthen the policy further

  **Important**:
  - You must always use the report_stance tool provided to analyze the evidence and determine a company's position on the provided specific climate policy query.
  - Do not respond without using the tool.
  - You can only give a score between -2 and 2.
  
  ---
  Question: Is the organization supporting policy and legislative measures to address climate change: Renewable energy legislation, targets, subsidies, and other policy?

  Context: As major businesses and employers in North Carolina, we are writing to you to express our support for the third-party leasing program in House Bill 589, Competitive Energy Solutions for NC, and to identify the Green Source Rider program as an area in need of further improvement during implementation. We applaud the numerous energy stakeholders and legislators who have worked to draft this consensus legislation over the past nine months, and we remain grateful to Speaker Tim Moore and Senate President Pro Tempore Phil Berger for convening the energy stakeholders’ process last September. We believe the Green Source Rider (GSR) provision in HB589 requires improvement to ensure customers will participate in the program. HB589 takes a few steps back from the previous pilot program, which only proved viable for three North Carolina businesses. We are concerned that preventing customers from achieving 100% renewable targets and by prescribing certain program requirements could negatively impact the viability of the GSR program. We believe that choice and competition in the renewable energy sector are as important as it is in all other aspects of our businesses and supply chains. More choices for companies to access renewable energy would give North Carolina businesses a competitive edge and allow us to keep our energy investment dollars here in the state. Establishing a cost-competitive corporate renewable purchasing  mechanism that works for diverse businesses, while ensuring no additional cost to non-participating customers, has been successfully achieved in 20 other states and many international markets-leading to over $15 billion in direct corporate investment.

  Stance: 2
  
  ---
  Question: Is the organization transparent and clear about its position on climate change science?

  Context: Humanity is consuming natural resources at an astonishing rate. During the 20th century, global raw material use rose at about twice the rate of population growth.  Every year, humanity consumes far more than what the planet can naturally replenish. In 2020, global demand for resources was 1.6 times what the earth's ecosystems can regenerate in a year. These statistics highlight the need to rethink the take-make-waste economic model--in which we take a natural resource, make a product from it or burn it for fuel, and eventually send what remains to the landfill as waste--that human societies have followed since the Industrial Revolution. The consequences of this model have contributed to significant global challenges such as climate change, extreme weather events, and plastic pollution.

  Stance: 2
  
  ---
  Question: Is the organization supporting the science-based response to climate change as set out by the IPCC?

Context: James Quincey, Chairman and CEO of The Coca-Cola Company, has joined heads of organizations that include L’Oreal, IKEA Foundation, World Wildlife Federation and the World Economic Forum, along with the Ellen MacArthur Foundation, in support of a circular economy. In a statement published in the Financial Times Weekend, they pledged to "build back better" after the challenges from the global coronavirus pandemic by designing out waste from their systems. "As the world faces unprecedented challenges," the statement said, "we are more committed than ever to accelerating the transition to a circular economy, creating solutions that combine economic opportunity with benefits to wider society and the environment."  letter: Many have already called for a response to the devastating impacts of this pandemic that does not turn attention away from other global challenges such as climate change and pollution. The circular economy offers a solution for how to do so.

  Stance: 1
  
  ---
  Question: Is the organization supporting an IPCC-aligned transition of the economy away from carbon-emitting technologies, including supporting relevant policy and legislative measures to enable this transition?

  Context: China should be Europe's role model when it comes to switching to green transport because Beijing is focusing on one technology  - batteries  - while the EU is mulling many more, the boss of MAN Truck & Bus Alexander Vlaskamp told POLITICO. Vlaskamp argued that Europe should adopt Beijing’s top-down approach that targets electrification only, which allowed China to become a front-runner in the electric vehicle (EV) sector, rather than spend time and money toying with other clean fuel alternatives. China has heavily subsidized the sector and "that is how they learned to build batteries," he told POLITICO and other journalists before this week’s European Parliament vote on the revision of CO2 truck standards. "If you want Europe to lead the CO2-neutral transport, then we have to invest in building battery plants, building battery electric trucks, building the grid," Vlaskamp said, adding that China has done that "orchestration." EU institutions are in agreement that new heavy-duty vehicles should cut their emissions by 90 percent by 2040, which means a revolution in how they’re powered. But there are strong lobbies in Europe  - especially in Germany  - for the combustion engine to be given another lease on life by allowing cars and trucks to use CO2-neutral fuels.  Vlaskamp expressed strong concern about the risks of investing in too many technologies, pointing at cities that have invested in hydrogen buses which are now parked thanks to high hydrogen prices. Looking for direction He added he is "disappointed" about the EU’s lack of clear guidance on clean fuels. The European Parliament on Tuesday approved an amendment allowing trucks using such fuels to be used by manufacturers to reach their climate targets. However, a proposal to include a carbon correction factor  - allowing truck makers to count fuels deemed CO2-neutral under the Renewable Energy Directive as emission savings for their fleets  - was rejected. It was heavily criticized by MAN and other businesses. MAN, a subsidiary of Volkswagen, is focusing primarily on electrifying its vehicles to hit its target of becoming carbon neutral by 2050.

  Stance: 1
  
  ---
  Question: To what extent does the organization express the need for regulatory intervention to resolve the climate crisis?

  Context: The window to avert irreversible and catastrophic climate change is closing rapidly. Global emissions must fall by about half by 2030 to meet the internationally agreed target of 1.5C of heating, but emission levels are still rising. Urgent and collective action from governments and businesses is needed to avoid the most severe climate impact.

  Stance: 0
  
  ---
  Question: Is the organization supporting an IPCC-aligned transition of the economy away from carbon-emitting technologies, including supporting relevant policy and legislative measures to enable this transition?

  Context: Biomethane trucks are essential for fossil-free transports.

  Stance: 0
  
  ---
  Question: Is the organisation transparent about its positions on climate change legislation/policy and its activities to influence it?

  Context: At Amazon, we are putting our scale and inventive culture to work on  sustainability not only because it is good for the environment, but also  for the customer. By diversifying our energy portfolio, we can keep  business costs low and pass along further savings to customers. It's a  win-win-win.  Clean Power Plan Amicus Brief In April 2016, Amazon joined Apple, Google, and Microsoft in filing a  legal brief that supports the continued implementation of the U.S.  Environmental Protection Agency's Clean Power Plan (CPP) and discusses  the technology industry's growing desire for affordable renewable energy  across the U.S. Read the brief here American Business Act on Climate Pledge In 2015, Amazon signed the White House's American Business Act on  Climate Pledge to express support for action on climate change and to  accelerate the transition to a low-carbon economy. The pledge brought  over 150 companies together to voice support for a strong outcome in the  2015 Paris climate negotiations and to demonstrate their ongoing  commitment to climate action.

  Stance: -1
  
  ---
  Question: Is the organization transparent about its involvement with industry associations that are influencing climate policy, including the extent to which it is aligned with these groups on climate?

  Context: Please enter the details of those trade associations that are likely to take a position on climate change legislation. Trade Association American Fuels and Petrochemical Manufacturers. your position on climate change consistent with theirs? Explain the trade association’s position AFPM members are strongly committed to clean air, water and waste reduction, have an outstanding record of compliance with the United States Environmental Protection Agency (EPA) and other regulators, and have invested hundreds of billions of dollars to dramatically reduce emissions as measured by EPA. have you, or are you attempting to, influence the position? (Chevron is a member of AFPM’s Board) We agree with the core message and commitment to clean air, water and waste reduction. Chevron shares the concerns of governments and the public about climate change risks and recognizes that the use of fossil fuels to meet the world’s energy needs contributes to the rising concentration of greenhouse gases (GHGs) in Earth’s atmosphere. GHGs contribute to an increase in global temperature. We apply cost-effective technologies to improve the energy efficiency of our base business operations and capital projects. As we work to address climate risks, we must create solutions that achieve environmental objectives without undermining growth of the global economy and our aspirations for a better quality of life for all.

  Stance: -1
  
  ---
  Question: Is the organization supporting policy and legislative measures to address climate change: standards, targets, and other regulatory measures directly targeting Greenhouse Gas emissions?

  Context: Peabody supports adoption of the Affordable Clean Energy (ACE) rule as a replacement for the Clean Power Plan as a significant step in ensuring reliable, resilient and affordable electricity across the U.S. ACE strengthens regulatory and investment certainty using a sensible and legally defensible "within the fenceline" approach to improve efficiencies of existing power plants. The proposed rule offers technically feasible and appropriate measures with the potential to deliver cost-effective, achievable and practical solutions for reducing emissions while minimizing disruptions to electricity generation. Furthermore, ACE offers individual states greater flexibility in the development and timing of state implementation plans, avoiding a "one-size-fits-all" approach to managing distinct and diverse needs. Peabody has long advocated for technology solutions to meet the three-part goal of energy security, economic progress and environmental solutions and we believe ACE satisfies these objectives. 

  Stance: -2
  
  ---
  Question: Is the organization supporting an IPCC-aligned transition of the economy away from carbon-emitting technologies, including supporting relevant policy and legislative measures to enable this transition?

  Context: The U.S. upstream industry is safely and sustainably delivering the energy the world needs, while also leading in the reduction of Green House Gas (GHG) emissions. We believe that the U.S. federal government and U.S. allies should strongly support continued development of top-quality U.S. oil and gas resources to help meet the current and future energy demands of the world.  

  Stance: -2
  
  ---
\end{lstlisting}

\section{Scores}
\label{appendix:b}
Scores have the following meaning according to LobbyMap's Methodology page -
\begin{itemize}
    \item -2: Position contradicts IPCC analysis OR Opposes policy
  \item  -1: Position appears misaligned with IPCC analysis OR Unsupportive of policy and/or communicates support for the policy but with major caveats and/or conditions that would weaken the strength of the proposal
   \item  0: Unclear if position is aligned with IPCC guidance OR Unclear if position is supportive of policy
  \item  +1: Broad alignment with IPCC analysis OR General or high-level support for the policy
 \item  +2: Detailed position that is aligned with IPCC analysis OR Strong Support for the policy and/or advocacy that would strengthen the policy further

\end{itemize}

\section{Full Query List}
\label{sec:queries}
The following queries were used to guide evidence extraction:
\begin{itemize}
\item Query 1: "Is the organization supporting policy and legislative measures to address climate change: energy efficiency policy, standards, and targets"
\item Query 2: "Is the organisation transparent about its positions on climate change legislation/policy and its activities to influence it?"
\item Query 3: "Is the organisation supporting policy and legislative measures to address climate change: carbon tax"
\item Query 4: "Is the organization transparent and clear about its position on climate change science?"
\item Query 5: "Is the organization supporting policy and legislative measures to address climate change: Standards, targets, and other regulatory measures directly targeting Greenhouse Gas emissions"
\item Query 6: "Is the organization transparent about its involvement with industry associations that are influencing climate policy, including the extent to which it is aligned with these groups on climate?"
\item  Query 7: "Is the organization supporting an IPCC-aligned transition of the economy away from carbon-emitting technologies, including supporting relevant policy and legislative measures to enable this transition?"
\item Query 8: "Is the organization supporting policy and legislative measures to address climate change: Renewable energy legislation, targets, subsidies, and other policy"
\item Query 9: "Is the organization supporting the science-based response to climate change as set out by the IPCC? "
\item Query 10: "Is the organization supporting the UN FCCC process on climate change?"
\item Query 11: "Is the organisation supporting policy and legislative measures to address climate change: emissions trading."
\item Query 12: "To what extent does the organization express the need for regulatory intervention to resolve the climate crisis?"
\item Query 13: "Is the organization supporting policy and legislative measures to enhance and protect ecosystems and land where carbon is being stored?"
\end{itemize}

\section{Longest Common Subsequence}
\begin{algorithm}[H]
\caption{Compute LCS Length}
\begin{algorithmic}[1]
\Function{LCS}{$X$, $Y$}
    \State $m \gets |X|$, $n \gets |Y|$
    \State Initialize $L[0..m][0..n] \gets 0$
    \For{$i = 1$ to $m$}
        \For{$j = 1$ to $n$}
            \If{$X[i{-}1] = Y[j{-}1]$}
                \State $L[i][j] \gets L[i{-}1][j{-}1] + 1$
            \Else
                \State $L[i][j] \gets \max(L[i{-}1][j], L[i][j{-}1])$
            \EndIf
        \EndFor
    \EndFor
    \State \Return $L[m][n]$
\EndFunction
\end{algorithmic}
\label{alg:lcs}
\end{algorithm}

\section{Dataset Structure}
\label{sec:data_structure}
Each data instance is a tuple:
(PDF, Company, Query, Evidence, Stance, Comment, Metadata)
where:

\begin{itemize}
    \item \textbf{PDF}: A company-related document (e.g., sustainability report, policy statement), often multilingual and diverse in layout.
    \item \textbf{Query}: One of 13 predefined climate policy questions representing lobbying subtopics (see Appendix \ref{sec:queries}).
    \item \textbf{Evidence}: A human-extracted text snippet relevant to the query, from the PDF.
    \item \textbf{Stance}: A discrete label from the set $\{-2, -1, 0, +1, +2\}$ representing opposition through support.
    \item \textbf{Comment}: A short justification written by an analyst explaining the stance assignment.
    \item \textbf{Metadata}: Includes the corporate entity, date, and region associated with the PDF
\end{itemize}

\section{Parsing and Chunking Configuration}
\label{appendix:parser_chunker_config}

\subsection*{Docling}
\label{appendix:docling_config}
The following settings were used for parsing all documents with the Docling toolkit:

\begin{itemize}
    \item \textbf{Device:} \texttt{mps} with 8 threads.
    \item \textbf{OCR:} Enabled using \texttt{easyocr} with:
    \begin{itemize}
        \item Full-page OCR forced
        \item Minimum confidence threshold: 0.5
    \end{itemize}
    \item \textbf{Table structure extraction:} Enabled with:
    \begin{itemize}
        \item Cell matching disabled
        \item TableFormer mode set to \texttt{accurate}
    \end{itemize}
\end{itemize}

\subsection*{PyMuPDF}
\label{appendix:pymupdf_config}
The PyMuPDF parser was used with the default settings.

\subsection*{Semantic Chunking}
\label{appendix:semantic_chunking}
The semantic chunking module was implemented using the Python library \texttt{CHONKIE}~\cite{chonkie2025}, and configured with the following options:

\begin{itemize}
    \item \textbf{Model:} \texttt{minishlab/potion-base-8M}
    \item \textbf{Similarity threshold:} 0.75
    \item \textbf{Double-pass merge:} Enabled
    \item \textbf{Chunk size:} 1536 tokens
    \item \textbf{Device:} \texttt{mps}
\end{itemize}

\subsection*{Layout Chunking}
\label{appendix:layout_chunking}
The layout chunking module was configured with the following options:

\begin{itemize}
    \item \textbf{Minimum chunk length:} 30 words.
    \item \textbf{Chunk boundary heuristic rules:}
    \begin{itemize}
        \item Headers are merged with the subsequent text block to preserve continuity.
        \item Subsequent headers are concatenated into a single header.
        \item Tables are treated as atomic units.
        \item Paragraphs shorter than the Minimum chunk length are merged with previous paragraphs.
        \item Paragraphs ending with colons are merged with previous paragraphs unless the latter begins with a header marker. 
    \end{itemize}
\end{itemize}

\section{LLM Evaluation}
\label{appendix:llm_eval}
The reported evaluations were performed on a sample of 2018 evidences extracted from 10 companies. This is an unbalanced evaluation set which is more representative of what an analyst might observe.

\begin{figure}
    \includegraphics[scale=0.2]{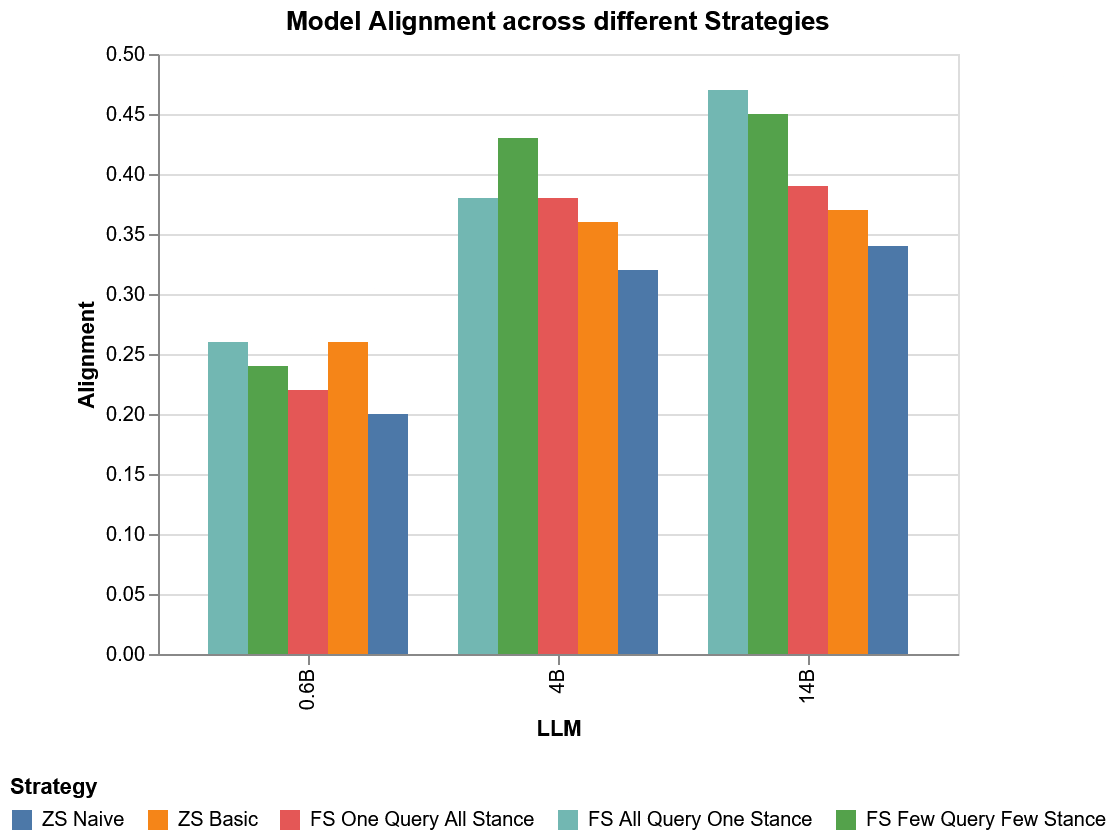}
    \caption{Model Alignment on balanced evaluation subset}
    \label{alignment_b}
\end{figure}

Figure \ref{alignment_b} shows the same evaluation on a balanced set of 1000 samples where for each stance we have 200 samples. Interestingly we observe similar results.

\section{When Retrieval Outperforms Ground Truth Results}
\begin{table}[ht]
\centering
\small
\renewcommand{\arraystretch}{1.2}
\begin{tabular}{lccc}
\toprule
\textbf{Metric} & \textbf{FR} & \textbf{AR} & \textbf{BM} \\
\midrule
Faithfulness      & 0.49 & 0.57 & 0.52 \\
Helpfulness       & 0.65 & 0.58 & 0.54 \\
Conciseness        & 0.77 & 0.82 & 0.88 \\
Outperform (\%)    & 46.2 & 30.8 & 23.1 \\
\bottomrule
\end{tabular}
\caption{Average oracle scores for retrieval strategies that outperformed Ground Truth in stance generation.}
\label{tab:oracle_outperforming}
\end{table}
\end{document}